%% file: main.tex
\documentclass[sigconf, nonacm]{acmart}

\AtBeginDocument{%
  \providecommand\BibTeX{{%
    Bib\TeX}}}

\setcopyright{acmlicensed}
\copyrightyear{2018}
\acmYear{2018}
\acmDOI{XXXXXXX.XXXXXXX}

\acmConference[Conference acronym 'XX]{Make sure to enter the correct
  conference title from your rights confirmation emai}{June 03--05,
  2018}{Woodstock, NY}
\acmISBN{978-1-4503-XXXX-X/18/06}

\usepackage{multirow}
\usepackage{caption}
\usepackage{subcaption}
\usepackage{multicol}
\usepackage{enumitem}

\usepackage{amssymb}
\usepackage{amsthm}
\usepackage{amsmath,amsfonts}
\usepackage{newtxmath}

\definecolor{background_gray}{gray}{0.9}
\usepackage{colortbl}
\input{macro}
\usepackage{float}

\usepackage{algpseudocode,algorithm}
\usepackage[algo2e,ruled,linesnumbered]{algorithm2e}

\SetCommentSty{mycommfont}

\let\oldnl\nl
\newcommand{\nonl}{\renewcommand{\nl}{\let\nl\oldnl}}

\usepackage{commath}

\def\BibTeX{{\rm B\kern-.05em{\sc i\kern-.025em b}\kern-.08em
    T\kern-.1667em\lower.7ex\hbox{E}\kern-.125emX}}

\begin{document}

\title{Distribution-aware Online Continual Learning for Urban Spatio-Temporal Forecasting
}

\author{Chengxin Wang}
\affiliation{%
  \institution{National University of Singapore}
  \country{Singapore}
}
\email{cwang@comp.nus.edu.sg}
\authornote{Work was done when the author interned at Comfort Transportation
Pte Ltd.
}

\author{Gary Tan}
\affiliation{%
  \institution{National University of Singapore}
  \country{Singapore}
}
\email{gtan@comp.nus.edu.sg}

\author{Swagato Barman Roy}
\affiliation{%
  \institution{CDG Zig Pte Ltd.}
  \country{Singapore}
}
\email{swagatobr@comfortdelgro.com}

\author{Beng Chin Ooi}
\affiliation{%
  \institution{National University of Singapore}
  \country{Singapore}
}
\email{ooibc@comp.nus.edu.sg}

\input{sections/0.abstract.tex}

\settopmatter{printacmref=false}

\maketitle

\input{sections/1.introduction}

\input{sections/2.formulation}

\input{sections/3.method}

\input{sections/4.experiments}

\input{sections/5.relatedwork}
\input{sections/6.conclusion}

\bibliographystyle{ACM-Reference-Format}
\balance
\bibliography{main}


\end{document}

%% file: macro.tex
\usepackage{color}
\usepackage{subcaption}
\usepackage{stmaryrd}
\usepackage{placeins}
\usepackage{flushend}   
\newcommand\name{\text{DOST}\xspace}

\makeatletter
\newcommand{\printfnsymbol}[1]{%
  \textsuperscript{\@fnsymbol{#1}}%
}
\makeatother

%% file: sections/0.abstract.tex
\begin{abstract}

Urban spatio-temporal (ST) forecasting is crucial for various urban applications such as intelligent scheduling and trip planning.
Previous studies focus on modeling ST correlations among urban locations in offline settings, which often neglect the non-stationary nature of urban ST data, particularly, distribution shifts over time.
This oversight can lead to degraded performance in real-world scenarios.
In this paper, we first analyze the distribution shifts in urban ST data, and then introduce \name,
a novel online continual learning framework tailored for ST data characteristics.
\name employs an adaptive ST network equipped with a \textit{variable-independent adapter} to address the unique distribution shifts at each urban location dynamically.
Further, to accommodate the gradual nature of these shifts, we also develop an \textit{awake-hibernate learning strategy} that intermittently fine-tunes the adapter during the online phase to reduce computational overhead. This strategy integrates a streaming memory update mechanism designed for urban ST sequential data, enabling effective network adaptation to new patterns while preventing catastrophic forgetting.
Experimental results confirm \name's superiority over state-of-the-art models on four real-world datasets, providing online forecasts within an average of 0.1 seconds and achieving a 12.89\% reduction in forecast errors compared to baseline models.
\end{abstract}

%% file: sections/1.introduction.tex
\section{Introduction}

Modern Intelligent Transportation Systems (ITS)~\cite{an2011survey,ran2012modeling} rely on extensive sensor networks deployed across urban areas to monitor traffic conditions.
These sensors produce vast amounts of spatio-temporal (ST) data that exhibit both spatial correlations and temporal dynamics.
Accurate forecasting of these urban ST data is crucial for smart city applications, such as intelligent scheduling~\cite{tong2017simpler,ji2020interpretable,han2024bigst}, traffic management\cite{ZhangHXXDBZ021,wu2021autocts,wang2023urban}, and trip planning~\cite{li2018multi,fu2019titan,fang2021mdtp}.

\input{figures/intro/data_distrubion}

Spatio-temporal (ST) correlations~\cite{jin2303spatio,duan2023localised,STMeta} play a critical role in urban forecasting, as traffic conditions at one location are influenced both by its historical data and by data from neighboring locations.
This has driven substantial efforts in developing advanced ST networks to capture urban ST dependencies~\cite{duan2023localised,han2023generic,10.14778/3551793.3551827}.
Nonetheless, most existing studies are conducted in offline settings with static data distributions, assuming stationary relationships between data over time.
In contrast, urban ST data are inherently dynamic with constantly evolving distributions, which renders offline models ineffective for practical deployment.
To address distribution shifts in streaming data, online continual learning has proven effective in forecasting tasks such as long-term time series forecasting (LTSF)~\cite{pham2022learning,zhang2023onenet}, natural language processing (NLP)~\cite{houlsby2019parameter,pfeiffer2020adapterhub}, and stock prediction~\cite{song2023follow,zhao2023doubleadapt}. However, these methods cannot be directly applied to urban ST forecasting for the following two reasons.

First, existing methods either immediately update upon receiving new data~\cite{pham2022learning,zhang2023onenet, houlsby2019parameter}, or delay updates until a new batch or domain is available~\cite{lopez2017gradient, cermelli2022incremental}.
Immediate updates are essential for tasks characterized by rapidly changing temporal patterns, such as Long-Term Series Forecasting (LTSF) and Natural Language Processing (NLP),
whereas batch updates are more common in scenarios that experience abrupt domain shifts, such as image classification.
However, urban ST data typically undergo gradual shifts over time due to static urban zoning and functionality~\cite{qian2015spatial,yu2019exploring}, rendering both immediate and batch updates impractical.
As illustrated in Figure~\ref{fig:intro}, taxi demand in a city like Chicago might appear stable over weeks, yet significant variations emerge over longer spans, such as months.
Specifically, the estimated distributions in regions, such as Region 23 and Region 64, demonstrate substantial similarity between two consecutive weeks, for example, from 08/01/2020 to 14/01/2020 and from 15/01/2020 to 21/01/2020.
This data characteristic necessitates a finer-grained update approach to avoid performance degradation and excessive computational costs associated with over-frequent or delayed model updates.

Second, previous efforts have encountered challenges in adapting to the varied distribution shifts across different urban locations during ST forecasting.
Each urban area exhibits distinct shift patterns due to location-specific factors such as urban functionality~\cite{qian2015spatial,yu2019exploring}.
As Figure~\ref{fig:intro} illustrates, the estimated data distributions in Region 23 (a residential area) and Region 64 (a transportation hub) differ significantly, and their shift patterns vary considerably over extended periods, for example, from 15/01/2020 to 31/03/2020.
Traditional adaptation strategies, such as model fine-tuning~\cite{pham2022learning,zhang2023onenet}, which updates all model parameters with the most recent data, or parameter-efficient tuning~\cite{houlsby2019parameter,pfeiffer2020adapterhub}, which selectively fine-tunes specific layers, can address these distribution shifts, but often at the cost of high computational demands and frequent unnecessary updates. 
Although recent advancements in parameter-efficient tuning~\cite{houlsby2019parameter,pfeiffer2020adapterhub,hu2021lora} and continuous learning~\cite{zhang2021dac, zhang2022urban,wang2023pattern} have shown promise in managing these shifts effectively, they also ignore the unique shifts at individual locations, thus falling short of the specific needs of urban ST forecasting.

\setlength{\emergencystretch}{3em}
To address the above challenges, we propose a novel \textbf{D}istribution-aware \textbf{O}nline continual learning framework for urban \textbf{S}patio-\textbf{T}emporal forecasting (\name) to handle distribution shifts in streaming urban ST data.
\name leverages inherent urban ST behaviors to enhance online urban forecasting, which considerably improves performance and efficiency.
In particular, to explicitly model the varying distribution shifts across urban locations over time, we introduce an adaptive ST network with a plug-and-play adapter named Variable-Independent Adapter (VIA).
VIA customizes adapters for respective urban locations, which effectively update the network in response to location-specific distribution shifts.
In addition, we introduce a novel Awake-Hibernate (AH) learning strategy to align network updates with the gradual shift characteristic of urban ST data.
This strategy alternates between awake and hibernate phases to minimize unnecessary updates and reduce computational costs.
During the awake phases, VIA is precisely fine-tuned to quickly adapt to new distribution patterns, while in the hibernate phases, all model parameters are frozen to conserve resources.
The network fine-tuning employs a Streaming Memory Update (SMU) mechanism, which adopts a small episodic memory selected from recent updates to ensure timely and effective adaptation without overfitting or catastrophic forgetting.

Our main contributions are summarized as follows:

\begin{itemize}[leftmargin=*]
    \item
    We propose \name, a novel distribution-aware online continual learning framework tailored for urban ST forecasting, which effectively balances between update phases to align with gradual urban distribution shifts, thus avoiding inefficient training cycles. 
    \item 
    We introduce the Variable-Independent Adapter VIA, which enables the network to adapt to diverse and evolving urban distribution shifts across different locations.
    \item
    Extensive experimental results confirm \name's superiority over state-of-the-art models on four real-world datasets, which delivers online forecasts in real-time, within only 0.1 seconds, and reduces forecast errors by 12.89\% compared to baseline models.
\end{itemize}

%% file: figures/intro/data_distrubion.tex
\begin{figure}[!t]
\vspace{1em}
\begin{center}
\includegraphics[width=.96\linewidth]{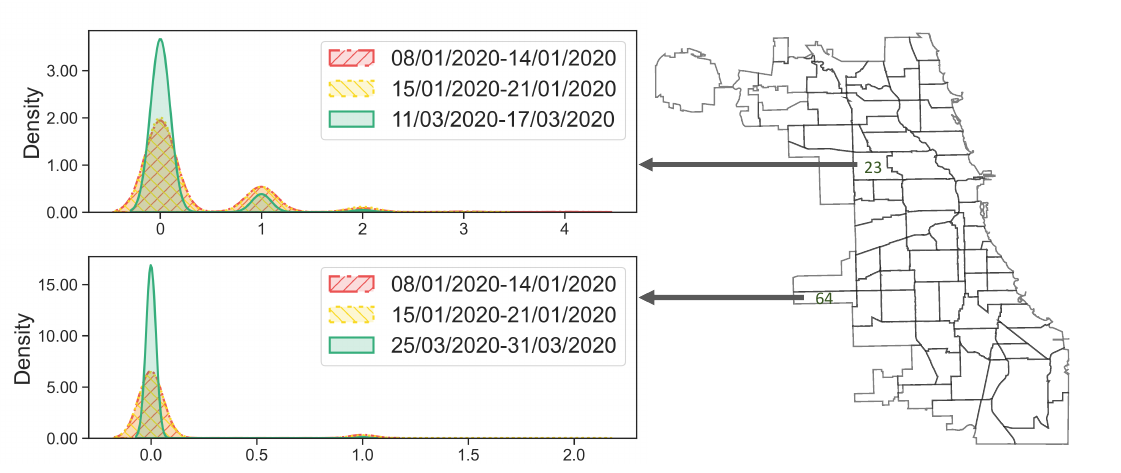}
\end{center}
\vspace{-0.5em}
\caption{
An illustration of Chicago's taxi demand distribution shift, estimated with Kernel Density Estimator (KDE).
Left hand side: estimated distributions for Region 23 (upper) and Region 64 (lower); right hand side: a visualization of Chicago's community regions. 
}
\vspace{-1em}
\label{fig:intro}
\end{figure}

%% file: sections/2.formulation.tex
\section{PRELIMINARIES} 

\input{figures/method/overview_online}

\vspace{0.5em}
\noindent \textbf{Definition 1 (Urban Spatio-Temporal Data)}:
Urban data, including region-based data (e.g., taxi demand and crowd flow) and road-based data (e.g., vehicle speed and traffic volume), are spatio-temporal (ST) data as they have both spatial and temporal dimensions. Region-based data covers \( N \) community-defined regions, while road-based data involves  \( N \) specific roads. Both data types evolve temporally within each region or road.

\vspace{0.5em}
\noindent \textbf{Definition 2 (External Factors)}:
Urban ST data often correlate with external factors, such as time of day and day of the week, as human activities and traffic patterns are shaped by daily routines and weekly cycles.

\begin{sloppypar}
\vspace{0.5em}
\noindent
\textbf{Online Urban Spatio-Temporal Forecasting:} In real-world applications, data arrives sequentially, denoted as ${X}_{\tau:\infty}$. The objective is to process this ongoing data stream and forecast future traffic conditions across urban locations for the next \( H \) time intervals at any given time step \( \tau \). In other words, given the past observed data within a look-back window \( L \), i.e., $\mathcal{X}_{\tau} = X_{\tau-L+1:\tau} = (X_{\tau-L+1}, X_{\tau-L+2}, \ldots, X_{\tau}) \in \mathbb{R}^{N \times L \times d}$, and the spatial adjacency matrix of urban locations $\mathbf{A} \in \mathbb{R}^{N \times N}$, the aim is to predict future urban ST conditions \(\hat{\mathcal{Y}}_{\tau}= \hat{X}_{\tau:\tau+H} = (\hat{X}_{\tau+1}, \hat{X}_{\tau+2}, \ldots, \hat{X}_{\tau+H}) \in \mathbb{R}^{N \times H \times d}\), where \( d \) represents the dimension of the data features.
\end{sloppypar}

%% file: figures/method/overview_online.tex
\begin{figure*}[!ht]
\begin{subfigure}[c]{.9\textwidth}
  \centering
  \includegraphics[height=2.6cm]{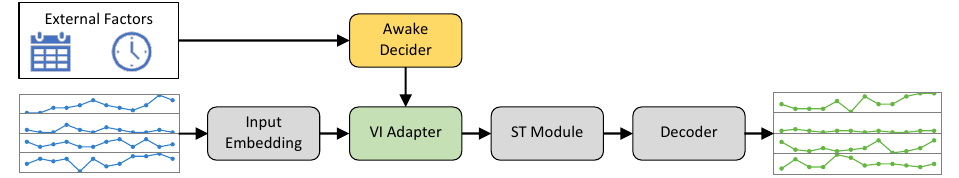}
  \vspace{-0.5em}
  \caption{Adaptive spatio-temporal network}
  \vspace{0.5em}
\end{subfigure}
\begin{subfigure}[c]{.9\textwidth}
  \centering
    \includegraphics[height=4cm]{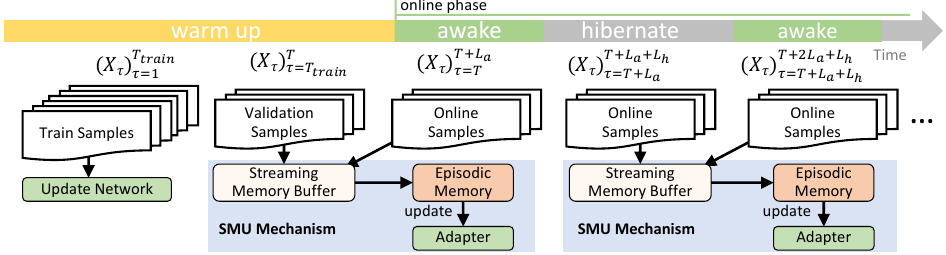}
  \caption{Awake-hibernate learning strategy}
\end{subfigure}
\caption{
Overview of \name, which employs two strategies: (a) Adaptive spatio-temporal (ST) network for online learning, where modules are represented in three colors: gray for traditional modules, green for the adapter, and yellow for the awake decider. 
(b) Awake-Hibernate (AH) learning strategy, which alternates network updates between awake and hibernate phases. During the awake phase, the adapter is fine-tuned using the Streaming Memory Update (SMU) mechanism, while during the hibernate phase, all parameters are frozen. Note: The Memory Placeholder in the SMU mechanism is omitted in this figure.}
\vspace{-1em}
\label{fig:overview}
\end{figure*}

%% file: sections/3.method.tex
\section{\name}
\label{sec:method}
In this section, we present our proposed model \name, as depicted in Figure~\ref{fig:overview}. 
\name employs two main strategies to tackle distribution shifts in sequential urban ST forecasting: an adaptive spatio-temporal network for online learning (Section~\ref{sec:ast}) and an awake-hibernate learning strategy for efficient model updates (Section~\ref{sec:ah}).

\subsection{Adaptive Spatio-Temporal Network}
\label{sec:ast}
Recognizing that future urban conditions are influenced by past ST correlations, various offline ST networks have been designed to capture these dependencies.
These networks typically utilize \textit{ST Modules} to capture spatial dependencies via graph neural networks (GNNs)~\cite{DBLP:STGCN, wu2020MTGNN} and attention mechanisms~\cite{zhou2020spatiotemporal,zhao2023spatio}, and exploit temporal dependencies with convolutional neural networks (CNNs)~\cite{graphwvenet, wu2020MTGNN}, recurrent mechanisms (RNNs)~\cite{DBLP:DCRNN, bai2020adaptive}, attentions~\cite{xu2020spatial, zhou2020spatiotemporal}, and multilayer perceptrons (MLP)~\cite{shao2022spatial,zhang2023mlpst}. However, these networks cannot adapt to location-specific time-evolving distributions in online urban ST forecasting.
To address this, we introduce an adaptive spatio-temporal network to learn online distribution shifts, as illustrated in Figure~\ref{fig:overview} (a). This network contains \textit{Traditional Modules} from offline ST networks to model the ST correlations, a \textit{Variable-Independent (VI) Adapter} to learn location-specific shifts, and an \textit{Awake Decider} to support the awake-hibernate learning strategy.

\subsubsection{\textbf{Traditional Modules}}
Traditional (i.e., offline) ST networks typically consist of three modules: \textit{Input Embedding}, \textit{Spatio-Temporal (ST) Module}, and \textit{Decoder}. In general, they begin with an \textit{Input Embedding} that transforms the raw inputs $\mathcal{X}_{\tau} \in \mathbb{R}^{N \times L \times d}$ into high-dimensional representations through a fully connected layer $FC(\cdot)$:

\begin{equation}
{\mathbf{h}}= FC(\mathcal{X}_{\tau}; \mathbf{W}_{e}),
\label{eq:emb}
\end{equation}
where $\mathbf{h} \in \mathbb{R}^{N \times L \times d_{h}}$, $d_{h}$ denotes the feature dimensions. 

Then, the spatio-temporal correlations can be captured via the \textit{ST Module}, which can be many existing ST networks, based on the input embedding $\mathbf{h}$ and the spatial adjacency matrix $\mathbf{A}$:

\begin{equation}
\tilde{\mathbf{h}}= {f}_{st}(\mathbf{h}, \mathbf{A}; \mathbf{W}_{st}),
\label{eq:st_net}
\end{equation}
where $\tilde{\mathbf{h}} \in \mathbb{R}^{N \times d_{o}}$ denotes the high-level ST representations, $d_{o}$ represents the feature dimensions, $\mathbf{W}_{st}$ are the learnable parameters, and ${f}_{st}$ denotes functions of the \textit{ST Module}. By default, we employ the ST network in GWNet~\cite{graphwvenet} as our \textit{ST Module}. Finally, future urban ST conditions can be predicted via a \textit{Decoder}, e.g., a fully connected layer:

\begin{equation}
\mathcal{\hat{Y}} = FC(\tilde{\mathbf{h}}; \mathbf{W}_{d}),
\label{eq:decoder}
\end{equation}
where $\mathcal{\hat{Y}} \in \mathbb{R}^{N \times H \times d}$, and $\mathbf{W}_{d}$ is the learnable parameters.

\subsubsection{\textbf{Variable-Independent Adapter (VIA)}}
Distribution shifts can vary significantly across different city locations. For example, data distribution in school regions may change dramatically during school holidays, whereas it remains stable in CBD regions.
To address these location-specific shifts, we propose a \textit{Variable-Independent Adapter} (VIA), as shown in Figure~\ref{fig:adaptor}. VIA consists of $N$ sub-adapters, each designed to handle the distribution shifts in a specific location, ensuring accurate adaptation without interference from irrelevant changes in other locations.

For each urban location $n$, VIA employs a sub-adapter to transform the original input embedding, learned from Equation~\ref{eq:emb}, into a location-specific adapted embedding:

\begin{equation}
{\mathcal{H}^{(n)}}= f_{a}({\mathbf{h}^{(n)}}; \mathbf{W}_{a}^{(n)})+ \mathbf{h}^{(n)} = \sigma\left( \mathbf{h}^{(n)} \mathbf{W}_{a_1}^{(n)}\right)\mathbf{W}_{a_2}^{(n)} + \mathbf{h}^{(n)},
\end{equation}
where $\mathcal{H}^{(n)} \in \mathbb{R}^{L \times d_{h}}$, $f_{a}$ represents the non-linear transformation function, i.e., multi-layer perceptron (MLP) layers, $\sigma$ refers to ReLU function, $\mathbf{W}_{a}^{(n)}$ is learnable parameters specific to location $n$, $\mathbf{W}_{a_1} \in \mathbb{R}^{d_{h} \times d_{m}}$, and $\mathbf{W}_{a_2} \in \mathbb{R}^{d_{m} \times d_{h}}$.
The final adapted features for all urban locations $\mathcal{H} \in \mathbb{R}^{N \times L \times d_{h}}$ can be obtained by concatenating $\mathcal{H}^{(n)}$ for each location $n$. The skip connection~\cite{he2016deep} is used to preserve the untransformed features. Thus, Equation~\ref{eq:st_net} in \name is updated to:

\begin{equation}
\tilde{\mathbf{h}} = f_{st}(\mathcal{H}, \mathbf{A}; \mathbf{W}_{st}),
\label{eq:st_net_v1}
\end{equation}
Note that VIA functions as a plug-and-play component before the variable mixing networks, i.e., \textit{ST Module}. This design is crucial due to the complex ST correlations in urban ST data, necessitating the variable mixing networks to model these dependencies~\cite{shao2023exploring}.

\input{figures/method/adapter}

\subsubsection{\textbf{Awake Decider}}\label{sec:awake_decider}
Urban ST data distribution typically remains stable over short periods, such as weeks, reducing the necessity for continuous network updates.
We introduce the \textit{Awake Decider} to alternate network updates between awake and hibernate phases, thus preventing unnecessary updates in real-time forecasting.
Recognizing weekly periodic patterns in urban ST data~\cite{shi2018discovering, wang2022periodic}, the \textit{Awake Decider} leverages external factors, i.e., date and time, to align the scheduling of awake and hibernate phases with the data's weekly patterns. Specifically, each awake-hibernate (AH) cycle consists of an \textit{awake phase} $L_{a}$ and a \textit{hibernate phase} $L_{h}$. Both phases are proportionate to the week's span $L_{w}$, with \( L_{h} = \lambda L_{a} \propto L_{w} \), where $\lambda$ is the AH parameter.

\subsection{Awake-Hibernate Learning Strategy}
\label{sec:ah}

Urban ST data distributions exhibit gradual shifts: they remain stable over short spans, such as consecutive weeks, but undergo significant changes over longer periods due to natural factors like seasonal variations that affect human activity patterns~\cite{cools2010assessing}.
Traditional offline ST networks, trained on static datasets, often struggle to adapt to these shifts, leading to less accurate predictions over extended inference periods. To address this gradual distribution shift, we introduce an \textit{Awake-Hibernate (AH) learning strategy}, depicted in Figure~\ref{fig:overview} (b). 
This strategy intermittently fine-tunes the adapter to align with the nature of urban ST distribution shifts, ensuring precise and timely forecasts over time.
Designed for online sequential data, our \textit{AH learning strategy} features a key update mechanism, i.e., the \textit{Streaming Memory Update (SMU) Mechanism}, and comprises three phases: \textit{Warm up}, \textit{Awake}, and \textit{Hibernate}.

\subsubsection{\textbf{Streaming Memory Update (SMU) Mechanism}}
\label{sec:smb}
The awake phase involves fine-tuning the network to adapt to new incoming patterns. However, continuously updating the network can cause catastrophic forgetting~\cite{lu2018learning, de2021continual}. For example, during a one-week update phase, the model could forget Monday's patterns by Sunday, despite daily variations. 
Recent studies~\cite{chaudhry2019tiny,lopez2017gradient} have shown that retaining a memory of previously trained samples can reduce forgetting and stabilize training. Inspired by these works, we design a \textit{Streaming Memory Update Mechanism}, as illustrated in Figure~\ref{fig:mp_kdd}. This mechanism fine-tunes the network with a tiny \textit{Episode Memory} (EM) for multi-step ahead forecasting in streaming data, which comprises three main components: a Memory Placeholder (MP) to track recent samples, a Streaming Memory Buffer (SMB) to store the most relevant samples, and a Streaming Memory Update (SMU) to update the adapters with the selected EM from the SMB.

\input{figures/method/mem_placeholder}

\vspace{0.5em}
\noindent
\textbf{Memory Placeholder} (MP) is designed to track recent samples for multi-step ahead predictions. 
At each time step, the network receives only the current observations, which means the ground truth for the current sample, i.e., $X_{\tau+1:\tau+H}$, is not immediately available.
The MP addresses this by maintaining both past and current observations, thus enabling the network to select the most recent samples without revisiting the data sequence. Specifically, upon receiving new data at each $\tau$, the MP discards the oldest data, $X_{\tau-L-H}$, and incorporates the new data $X_{\tau}$, thus updating the MP to hold the most recent observations as $MP = X_{\tau-L-H+1:\tau} \in \mathbb{R}^{(L+H) \times N \times d}$ at $\tau$. 
The MP then extracts recent observations $MP_{\tau}^{x} = X_{\tau-L-H+1:\tau-H} \in \mathbb{R}^{L \times N \times d}$ and ground truths $MP_{\tau}^{y} = X_{\tau-H+1:\tau} \in \mathbb{R}^{H \times N \times d}$ to serve as inputs for the Streaming Memory Buffer.

\vspace{0.5em}
\noindent

\textbf{Streaming Memory Buffer} (SMB) selectively stores the most relevant samples. We consider samples from the latest AH cycle as the most relevant for two reasons inherent to urban ST data: (1) distant past data can become irrelevant for future predictions due to evolving patterns; (2) recurrent patterns often emerge within a single AH cycle due to gradual shifts and weekly periodicity. Therefore, we design a SMB $\mathcal{M}$ with $M$ memory slots to selectively retain observations and ground truths from the most recent hibernate phase and the current awake phase up to time $\tau$. At each time step $\tau$, given the recent observations $MP_{\tau}^{x}$ and ground truths $MP_{\tau}^{y}$ from the MP, $\mathcal{M}$ is updated via reservoir sampling~\cite{vitter1985random}. The probability of storing a sample in the SMB is $p = M/L_{ah}$, where $L_{ah} = L_{h} + L_{a}$ is the total duration of an AH cycle. We reset $\mathcal{M}$ at the start of each hibernate phase to ensure that the SMB only stores the most relevant samples. 
Thus, $\mathcal{M}$ at time $\tau$ can be represented as:

\vspace{-0.5em}
\begin{equation}
\small{
    \mathcal{M}_\tau = \begin{cases} 
    \{(x, y) \in (MP_{\tau}^{x}, MP_{\tau}^{y}) \mid \text{sampled with } p\} & \text{if } \tau \not\equiv 0 \pmod{L_{ah}}, \\
    \emptyset & \text{otherwise}.
    \end{cases}}
\end{equation}
\vspace{-0.5em}

\vspace{0.5em}
\noindent
\textbf{Streaming Memory Update} (SMU) efficiently updates the model with newly emerging patterns from incoming data while preventing catastrophic forgetting. It is tailored for urban ST streaming, differing from previous methods in three key aspects: (1) SMU selects episodic memory from the most relevant samples, instead of randomly selecting episodic memory from all past observations~\cite{chaudhry2019tiny, zinkevich2003online}; (2) rather than updating the network based on the very latest sample~\cite{zhang2023onenet, pham2022learning}, SMU does not explicitly incorporate the very latest sample for network updates; (3) SMU updates the network only during the awake phase, instead of fine-tuning the network immediately upon receiving new data~\cite{douillard2021plop, cermelli2022incremental}.

At each \textit{awake phase} time step, we employ SMU to fine-tune the adapter with a tiny \textit{Episodic Memory} (EM), which is selected from the SMB. Specifically, at each time step $\tau$, given the updated SMB $\mathcal{M}$, we first randomly select a tiny EM $\mathcal{M}_{e}$ of size $M_{e}$ from $\mathcal{M}$, where $M_{e} << M$. The selected EM $\mathcal{M}_{e}$ only includes data samples from the most recent AH cycle up to the current time $\tau$, serving to: (1) incorporate recent patterns from the latest AH cycle to prevent catastrophic forgetting, and (2) introduce randomness through non-sequential samples to prevent overfitting. The adapter, i.e., VIA, is then optimized based on errors for the selected $M_{e}$, followed by generating forecasts for $\tau$.
Note that $\mathcal{M}_{e}$ might not contain the very latest samples, i.e., $\mathcal{X}_{\tau-H}$ and $\mathcal{Y}_{\tau-H}$ as random sampling from $\mathcal{M}$ is adopted. However, this is sufficient to capture recent patterns, as distribution shifts within an AH cycle are generally stable.

\noindent
\subsubsection{\textbf{Warm Up Phase}} 
In real-world scenarios, data typically arrive in a sequential order, introduce challenges for model convergence due to limited randomness~\cite{nguyen2022globally}. To address this, during the \textit{warm up phase}, we divide historical data into training and validation sets. We shuffle the training set to introduce randomness, aiding in preventing overfitting, while the validation set remains unshuffled. In the validation phase, the samples are updated to the SMB $\mathcal{M}$ using reservoir sampling~\cite{vitter1985random}, preparing the model for the first online \textit{awake phase}. 
To this end, this \textit{warm up phase} can ensure the effective handling of upcoming data streams.

\subsubsection{\textbf{Awake Phase and Hibernate Phase}}
\name operates in alternating awake and hibernate phases during the online phase. The transition between these phases is controlled by the \textit{Awake Decider} (see Section~\ref{sec:awake_decider}). During the \textit{\textbf{Awake Phase}}, updates are applied to both the SMB and the model. Specifically, the SMB $\mathcal{M}$ is updated with newly received samples, and the network is fine-tuned using the SMU mechanism to adapt to new patterns. For computational efficiency, only the adapters are fine-tuned, while traditional modules remain frozen. In contrast, the \textit{\textbf{Hibernate Phase}} focuses solely on updating the SMB, suspending model updates due to the stability of short-term shifts to save computational costs. $\mathcal{M}$ is reset at the onset of this phase and then updated with new samples, preparing it for the upcoming AH cycle.

\input{algorithm/online_short_term}

\subsubsection{\textbf{Algorithm}}
Algorithm~\ref{alg:online} outlines the proposed \textit{AH learning strategy} for online urban ST forecasting. Specifically, given the incoming data stream, the algorithm operates at each timestep. At each timestep $\tau$, $MP \in \mathbb{R}^{(L+H)\times N \times d}$ stores observed samples $(X_{\tau-L-H+1}, \cdots, X_{\tau-1}, X_{\tau})$, allowing $\mathcal{M}$ to incorporate the most recent observations $MP^x$ and ground truths $MP^y$ for the time step $\tau-H$ via reservoir sampling. During the \textit{awake phase}, the SMU is executed to update the VIA with a small, randomly selected subset of memory from the SMB. At the beginning of each \textit{hibernate phase}, the SMB is reset. The algorithm involves both awake and hibernate phases during the online phase, and the time complexity differs between these phases. Considering the execution time of $f(\cdot)$ as $T_{f}$, during the \textit{awake phase}, the time complexity to update the SMB with the current observation is $O(NL + NH + M)$. The time complexity for the streaming memory update is $O(M_{e}T_{f})$, and for forecasting, it is $O(T_{f})$. Since $(NL + NH + M) \ll M_{e}T_{f}$, the total time complexity during the \textit{awake phase} is $O(M_{e}T_{f})$. During the \textit{hibernate phase}, the time complexity is $O(T_{f})$.

\subsection{Optimization}
Following the prior works~\cite{li2018multi, yao2018deep}, we adopt the Mean Absolute Error (MAE) as our loss function. The loss during the warm up phase is formulated as follows:
\begin{equation}
\mathcal{L}(\theta_{t}, \theta_{a}) = MAE(\hat{\mathcal{Y}}, \mathcal{Y}),
\end{equation}
where $\hat{\mathcal{Y}}$ represents the predicted ST conditions, $\mathcal{Y}$ denotes the actual ST conditions, and \( \theta_{t} \) and \( \theta_{a} \) refer to the sets of learnable parameters of the traditional modules and adaptive modules, respectively. 
For online adaptation, the loss during the awake phase is calculated as MAE based on the samples in the \textit{Episodic Memory}:
\begin{equation}
\mathcal{L}(\theta_{a}) = MAE(\hat{\mathcal{Y}}, \mathcal{Y}).
\end{equation}
\noindent
With this loss function, we can then train \name in an end-to-end manner effectively via popular gradient-based optimizers such as Adam~\cite{KingmaB14}, AdamW~\cite{loshchilov2017decoupled} and etc.

%% file: figures/method/adapter.tex
\begin{figure}[!t]
  \centering
    \includegraphics[height=2.4cm]{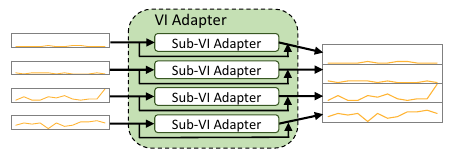}
\caption{
Overview of Variable-Independent Adapter (VIA), which comprises a series of Sub-VIAs, each dedicated to learning the distinct distribution shifts at specific urban locations.
}
\vspace{-1em}
\label{fig:adaptor}
\end{figure}

%% file: figures/method/mem_placeholder.tex
\begin{figure}[!ht]
\centering
  \includegraphics[width=1.0\linewidth]{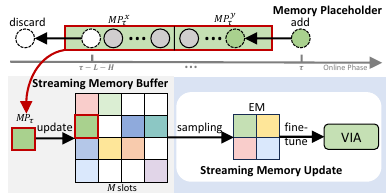}
\vspace{-1em}
\caption{
An illustration of the Streaming Memory Update (SMU) Mechanism. 
}
\vspace{-1em}
\label{fig:mp_kdd}
\end{figure}

%% file: algorithm/online_short_term.tex
\begin{algorithm2e}[t]
\scriptsize 
\SetAlgoNoEnd 
\nonl \textbf{Input:} 
Network $f(\cdot)$ learned during the warm-up phase, includes parameters $\theta_{t}$ for traditional modules and $\theta_{a}$ for the adapter; Validation dataset $\mathcal{D}_{val}$;
The length of the look-back window $L$ and prediction horizon $H$; The length of the awake and AH period $L_{a}$ and $L_{ah}$; The memory placeholder $MP$;
Online data stream $[{X}_{\tau}, {X}_{\tau+1}, \cdots,{X}_{\infty}]$.
\\
\nonl \hrulefill\\
$\mathcal{M} \leftarrow \emptyset $  \tcp*[l]{set the SMB to empty}
$awake \leftarrow true$ \tcp*[l]{set the online update to awake}

\tcp{Update the SMB with validation data} 
\ForEach{${(\mathcal{X}_{val}, \mathcal{Y}_{val})} \in \mathcal{D}_{val}$}{
   
     $\mathcal{M} \leftarrow \mathcal{M} \cup \{(\mathcal{X}_{val}, \mathcal{Y}_{val})\}$ 
     \tcp*[l]{reservoir sampling update} 
}

\tcp{Online phase} 
\ForEach{$\tau \in [0, \infty)$}{

\tcp{Update the SMB with current observations} 
$MP \leftarrow {X}_{\tau}$ \tcp*[l]{update memory placeholder}
 \If{$\tau >= H$}{
${MP}^{x}_{\tau} = X_{\tau-L-H:\tau-H}$\\
${MP}^{y}_{\tau} = X_{\tau-H+1:\tau}$\\
 $\mathcal{M} \leftarrow \mathcal{M} \cup \{({MP}^{x}_{\tau}, MP^{y}_{\tau})\}$  \tcp*[l]{reservoir sampling update} 
 }

\tcp{streaming memory update}
\If{$awake$}{
    Sample a tiny $\mathcal{M}_{e}$ from the $\mathcal{M}$;\\
    \ForEach{$X_{e} \in \mathcal{M}_{e}$}{
        $\hat{\mathcal{Y}}_{e} = f(\mathcal{X}_{e})$;
    }
    $\theta_{a} \leftarrow \mathcal{L}(\mathcal{\hat{Y}}_{e}, \mathcal{{Y}}_{e})$\\ 
}

$\hat{\mathcal{Y}}_{\tau} = \hat{\mathcal{X}}_{\tau+1:\tau+H} = f({\mathcal{X}_{\tau-L:\tau}})$;\\

\tcp{decide awake or hibernate for next time step} 
\uIf{$ \tau \quad \% \quad L_{a} == 0$ and  $\quad awake \quad$}{
    $awake \leftarrow false$\\
    $\mathcal{M} \leftarrow \emptyset $ \tcp*[l]{set the memory buffer to empty}
    }

\uElseIf{$ \tau \quad \% \quad L_{ah} == 0$ and  $\quad not \quad awake \quad$}{
    $awake \leftarrow true$\\
    }
}
\caption{Online Urban Spatio-Temporal Forecasting}
\label{alg:online}
\end{algorithm2e}

%% file: sections/4.experiments.tex
\section{Experiments}
\label{sec:evaluation}

\subsection{Experimental Settings}
\subsubsection{\textbf{Datasets}} 
\label{sec:dataset}

We evaluate \name on four real-world datasets, i.e., Chicago-T\footnote{https://data.cityofchicago.org/}, Singapore-T\footnote{https://www.cdgtaxi.com.sg/}, METR-LA~\cite{DBLP:DCRNN} and PEMS-BAY~\cite{DBLP:DCRNN}. 
Chicago-T and Singapore-T are region-based datasets, while METR-LA and PEMS-BAY are road-based datasets.
Detailed statistics of datasets are described in Table~\ref{dataset}.

\input{tables/datasets}

\input{tables/forecast_results}

\subsubsection{\textbf{Evaluation Metrics}} 

We follow the previous studies~\cite{wang2021libcity,shao2023exploring} to evaluate our model performance using three metrics: \textbf{Mean Absolute Error} (MAE), \textbf{Root Mean Squared Errors} (RMSE) and \textbf{Weighted Mean Absolute Percentage Error} (WMAPE).

\subsubsection{\textbf{Implementation Details}}
\name is trained on an NVIDIA GeForce RTX 3090 GPU using AdamW optimizer~\cite{loshchilov2017decoupled} with a learning rate of 0.001. We adopt an early-stop strategy, setting a patience parameter of 10 and a maximum number of epochs of 150 for all experiments. 
We set the look-back window $L$ to 12, forecast horizon $H$ to 12, AH parameter $\lambda$ to 1, SMB slot $M$ to 1000, and EM size $M_{e}$ to 8. 
For the dimensions, $d_{h}=32$, $d_{o} = 256$, and $d_{m} = 4$. $L_{a}$ is set to the total time intervals for a week, i.e., 672 for Chicago-T and Singapore-T datasets, and 2016 for METR-LA and PEMS-BAY datasets.
The data is divided into warm up and online phases in a 2:6 ratio. The warm up phase is further partitioned into a 4:1 ratio for training and validation. 

\subsubsection{\textbf{Baselines}}
\label{sec:baseline}
We compare \name against several widely used baselines for urban ST forecasting and Long-term Time Series Forecasting (LTSF).
HA~\cite{brockwell2016introduction} is a classical method. 
We also evaluate six strong models specifically for urban ST forecasting, i.e., STGCN~\cite{DBLP:STGCN}, GWNET~\cite{graphwvenet}, AGCRN~\cite{bai2020adaptive}, MTGNN~\cite{wu2020MTGNN}, GMSDR~\cite{liu2022msdr},
and PDFormer~\cite{pdformer}. Additionally, we examine six state-of-the-art LTSF methods capable of predicting traffic conditions, though not specifically tailored for short-term urban ST forecasting. These methods include 
REVIN~\cite{kim2021reversible}, PatchTST~\cite{nie2022time}, Dlinear~\cite{zeng2023transformers},
OnlineTCN~\cite{zinkevich2003online}, FSNet~\cite{pham2022learning}, and OneNet~\cite{zhang2023onenet}. 
Among them, OnlineTCN, FSNet, and OneNet are designed for online forecasting.

\subsection{Experimental Results \& Analysis}

\subsubsection{\textbf{Performance Comparison}}
Table~\ref{tab:taxi_demand} presents the prediction results of various baseline models and \name across four datasets. 
The results indicate that: 
(1) Models with advanced ST networks yield better results on most datasets, e.g., Singapore-T, METR-LA, and PEMS-BAY. This is because the significant spatial indistinguishability in urban ST data~\cite{shao2023exploring} requires advanced ST models to capture complex correlations. 
LTSF methods, which focus mainly on temporal correlations, struggle to learn these complex ST patterns.
(2) Online forecasting models, e.g., FSNet and \name, excel on datasets like Chicago-T, which have extended test phases. Conversely, methods designed to address out-of-distribution issues, e.g., REVIN and PatchTST, struggle with long-term distribution changes because they learn from fixed samples.
This highlights the effectiveness of online settings in real-world urban ST forecasting, where the ability to continuously adapt to new data patterns is crucial during prolonged testing phases.
(3) \name, tailored for urban ST forecasting, significantly outperforms all baseline models across various datasets. Unlike traditional ST networks, \name tackles distribution shifts via the AH learning strategy. It also addresses complex spatial correlations and location-specific shifts through its ST Modules and VIA, distinguishing it from general online learning methods. \name reduces MAE by an average of 12.89\% compared to baseline models across various datasets. T-test results across the four datasets confirm \name's consistent superiority over the leading baselines.

\input{tables/strategy}

\subsubsection{\textbf{Evaluating Strategies in \name}}

The two strategies in \name, i.e., adaptive ST network and AH learning strategy, can be seamlessly integrated with many existing offline urban ST forecasting methods. 
Table~\ref{tab:strategy} demonstrates the effectiveness of our strategies within various baselines, including STGCN, MTGNN, and GWNET.
The results indicate that: (1) Models enhanced with the adaptive ST network, i.e., STGCN*, MTGNN*, and GWNET*, surpass their original versions, thanks to the VIA that captures unique behaviors across various urban locations.
Although this enhancement requires additional parameters, it significantly boosts performance with an acceptable increase in parameters.
(2) Integrating the AH learning strategy into STGCN*, MTGNN*, and GWNET* further enhances forecasting performance by enabling the models to address distribution shifts at each urban location over time. This underscores the importance of AH learning in urban ST forecasting, even for short-span datasets with fewer distribution shifts like Singapore-T.
(3) Models integrating both strategies show substantial improvements over their base models. These strategies are plug-and-play options for many existing ST models, making the framework suitable for various urban scenarios.

\subsubsection{\textbf{Speed Comparison and Memory Usage}}

Table~\ref{tab:speed_mem_comparison} presents the inference time and memory usage of \name with various baselines on the Singapore-T dataset, which contains 13,093 test samples. Results are reported on a GTX 3090 GPU with 24,268 MB of memory.

\input{tables/test_mem}

The results indicate that: (1)
Offline methods such as STGCN, GWNET, MTGNN, and PatchTST have faster inference times than advanced online methods like FSNet and OneNet because they do not require backpropagation during the online phase.
(2) \name achieves the best performance with reasonable computational costs, outperforming advanced online methods by updating only the adapter rather than all parameters. 
It requires \(0.0266 \, \text{s}\) per sample during the hibernate phase and \(0.1099 \, \text{s}\) in the awake phase for both fine-tuning and forecasting.
This efficiency confirms \name's effectiveness in real-world urban ST forecasting.
(3) \name excels in performance with manageable memory usage. 
It employs the GWNET backbone to capture complex ST correlations, thus requiring more memory than FSNet and OneNet. 
However, by fine-tuning only the adapter, \name avoids excessive memory usage for gradient storage. Even though SMB needs memory to store past observations, with reasonable sizes for SMB $M$, spatial size $N$, look-back $L$, and prediction horizon $H$, the space complexity of SMB $O(MN(L+H))$ does not significantly increase overall memory usage.

\subsection{Ablation Study}
\label{sec:ablation}

Figure~\ref{fig:ablation} illustrates the effectiveness of each component in \name. 
\textbf{w/o VIA} denotes \name without the VIA, updating the default network with the AH strategy;
\textbf{w/o AH} refers to \name without the AH learning strategy, using only the adaptive ST network; 
\textbf{w/o VIAH} refers to \name without both VIA and AH.
\textbf{w VSA} replaces VIA with vanilla MLP layers and updates the network using AH.
\textbf{w/o Reset} does not reset the SMB at hibernate phase start;
\textbf{w/o SMU} refers to \name without SMU mechanism, instead it adopts the online update strategy from existing works~\cite{pham2022learning}, updating the model directly with the latest observations.

The results indicate that: 
(1) \textbf{w/o VIA} and \textbf{w/o AH} outperform \textbf{w/o VIAH}, demonstrating the effectiveness of our strategies for urban ST forecasting in an online setting.
(2) \textbf{w VSA} exhibits a performance decline, indicating that updating the network without considering location-specific distributions is insufficient. This proves the importance of VIA, which can adapt to various distribution shifts across locations over time.
(3) The inferior results of the \textbf{w/o Reset} verify our presupposition that samples from the distant past are not relevant to current forecasting.
(4) Increased errors in \textbf{w/o SMU} highlight the benefits of our SMU as it leverages historical knowledge to mitigate catastrophic forgetting and introduces randomness to avoid overfitting.

\input{figures/exp/ablation}

\subsection{Study on AH Learning Strategy}

Table~\ref{tab:online_update} presents the prediction results and the total inference time (in seconds) of our model using various online continual learning strategies, including:
\textbf{w/o H} omits the hibernate phase, updating the model at every time step.
\textbf{w ER} adopts the learning strategy from ER~\cite{chaudhry2019tiny}, utilizing a memory buffer and current observations.
\textbf{w ERH} extends \textbf{w ER} by including the hibernate phase.
\textbf{w SMUR} adopts the SMU mechanism and adds the most recent samples to the episodic memory.
\textbf{Full} fine-tunes all network parameters.

\input{tables/onlinestrategy}

The results indicate that:
(1) Omitting hibernate phases not only introduces computational costs but also leads to performance degradation, likely due to overfitting caused by too frequent updates. Additionally, the superior performance of \textbf{w ERH} over \textbf{w ER} demonstrates the beneficial role of hibernate phases in enhancing online learning for urban ST forecasting.
(2) \name performs comparably to \textbf{w SMUR}, indicating that random episodic memory sampling is adequate given the stable distribution shift within each AH cycle. Furthermore, \name outpaces \textbf{w SMUR} in speed as it does not require the concatenation of recent samples to episodic memory.
(3) More parameter updates lead to longer inference times. While the \textbf{Full} strategy achieves better results, it also requires higher computational resources. Notably, updating only VIA yields results comparable to \textbf{Full} but with a reduced computational load. Thus, we opt to update only the adapter during the online phases.

\subsection{Effects of Hyperparameters}

In Figure~\ref{fig:hyperparameter}, we study the effects of hyperparameters in \name on the PEMS-BAY dataset. 
The results indicate that: (1) Increasing $d_{m}$ from 4 to 16 lowers MAE and RMSE due to improved network capability. However, $d_{m}=8$ requires 128K more parameters compared to $d_{m}=4$. Since our model performs well with $d_{m}=4$, we select it as our default setting.
(2) \name performs best at $\lambda = 1$, with performance decreasing at higher $\lambda$ values, suggesting that the data distribution changes gradually over time. Eliminating either the awake phase ($\lambda = 0$) or the hibernate phase ($\lambda = \varnothing$) results in lower performance, highlighting the need for intermittent updates as the data distribution evolves. Surprisingly, even with an extended hibernate phase ($\lambda = 2$), \name performs better than without any hibernate phase, indicating that too frequent updates can cause overfitting and catastrophic forgetting.
(3) A larger SMB size ($M$) helps reduce MAE and RMSE, as it provides more historical pattern knowledge. However, an excessively large $M$ is unnecessary, as the SMB is reset at the end of each AH cycle.
(4) The absence of episodic memory, i.e., \(M_{e} = 0\), greatly diminishes performance. Conversely, small episodic memory sizes, i.e., $M_{e}=8$, are sufficient to introduce historical patterns and randomness, which helps mitigate overfitting to recent data and thus improves results.

\input{figures/exp/hyperparameter}

%% file: tables/datasets.tex
\begin{table}[!htbp]
\centering
\small
\vspace{-0.5em}
\tabcolsep=0.6mm
\caption{Statistics of the datasets.}
\vspace{-0.5em}
\label{dataset}
\begin{tabular}{l|l|l|l|l}
\hline
\textbf{Dataset} & \textbf{Chicago-T} & \textbf{Singapore-T} & \textbf{METR-LA} & \textbf{PEMS-BAY} \\ 
\hline
\textbf{Data Type} & Taxi Demand & Taxi Demand & Traffic Speed & Traffic Speed \\
\textbf{Time Span} & 01/01/2020 -  
&  06/02/2023 - 
& 01/03/2012 - 
& 01/01/2017 - \\
(dd/mm/yyyy) 
& 31/12/2023
& 06/08/2023 
& 27/06/2012
& 30/06/2017\\
\textbf{Time Interval} & 15 minutes & 15 minutes & 5 minutes & 5 minutes\\
\textbf{Spatial Size} & 77 & 87 & 207 & 325 \\
\hline
\end{tabular}
\vspace{-1em}
\end{table}

%% file: tables/forecast_results.tex
\begin{table*}[ht!]
\centering
\small
\caption{=Performance comparisons. The best results are bolded, and the most competitive results are underlined. Symbol $^\dagger$ and $^\ddagger$ indicate that \name achieves significant improvements with p < 0.001 and p < 0.05 over the most competitive results, respectively.
Experiments are repeated five times with different seeds on a GTX 3090 GPU. OOM denotes out-of-memory issues.}
\vspace{-0.5mm}
\tabcolsep=0.7mm
\label{tab:taxi_demand}
\begin{tabular}{l|ccc|ccc|ccc|ccc}
\hline
\multirow{2}{*}{\textbf{Method}} & \multicolumn{3}{c|}{\textbf{Chicago-T}} & \multicolumn{3}{c|}{\textbf{Singapore-T}} & \multicolumn{3}{c|}{\textbf{METR-LA}} & \multicolumn{3}{c}{\textbf{PEMS-BAY}}\\\cline{2-13}
& MAE$\downarrow$ & RMSE$\downarrow$ & WMAPE $\downarrow$ &MAE$\downarrow$ & RMSE$\downarrow$ &  WMAPE $\downarrow$ & MAE$\downarrow$ & RMSE$\downarrow$ & WMAPE $\downarrow$ &MAE$\downarrow$ & RMSE$\downarrow$ &  WMAPE $\downarrow$\\
\hline
HA  &  1.42 &  5.83 &  77.18\% 
& 12.91 & 29.67 & 71.22\% 
&38.27& 40.51& 68.19\% 
& 41.82& 42.50 & 66.33\%
\\
STGCN & 0.88\tiny{$\pm 0.01$}  & 2.84\tiny{$\pm 0.04$} & 45.04\%\tiny{$\pm 0.38\%$}  
& 8.23\tiny{$\pm 0.08$} & 15.50\tiny{$\pm 0.23$} & 46.17\%\tiny{$\pm 0.46\%$}
&\underline{4.52}\tiny{$\pm 0.02$} &\underline{8.41}\tiny{$\pm 0.08$}&\underline{8.34\%}\tiny{$\pm 0.05\%$}  
&1.93\tiny{$\pm 0.02$} &3.54\tiny{$\pm 0.03$} &3.07\%\tiny{$\pm 0.04\%$}\\ 
GWNET & 0.88\tiny{$\pm 0.03$} & 2.84\tiny{$\pm 0.16$} & 45.28\%\tiny{$\pm 1.41\%$} 
&8.43\tiny{$\pm 0.07$} & 15.70\tiny{$\pm 0.12$} & 47.28\%\tiny{$\pm 0.37\%$} 
& 4.57\tiny{$\pm 0.02$}& 8.51\tiny{$\pm 0.04$} &  8.46\%\tiny{$\pm 0.04\%$} 
&\underline{1.83}\tiny{$\pm 0.01$} & 3.50\tiny{$\pm 0.02$} &\underline{2.93\%}\tiny{$\pm 0.01\%$}\\ 
AGCRN & 0.84\tiny{$\pm 0.01$} &  2.60\tiny{$\pm 0.04$} &   42.98\%\tiny{$\pm 0.33\%$}  & 
\underline{8.13}\tiny{$\pm 0.03$} & \underline{15.29}\tiny{$\pm 0.08$} & \underline{45.47\%}\tiny{$\pm 0.15\%$}
&4.72\tiny{$\pm 0.01$}  &8.61\tiny{$\pm 0.03$} &8.87\%\tiny{$\pm 0.02\%$}  
&1.84\tiny{$\pm 0.02$}  &\underline{3.48}\tiny{$\pm 0.02$} &2.93\%\tiny{$\pm 0.03\%$} 
 \\
MTGNN & 0.90\tiny{$\pm 0.00$} & 2.87\tiny{$\pm 0.02$} & 46.14\%\tiny{$\pm 0.24\%$}
& 8.62\tiny{$\pm 0.07$} & 15.70\tiny{$\pm 0.12$} & 47.28\%\tiny{$\pm 0.38\%$} 
&4.63\tiny{$\pm 0.00$} &8.64\tiny{$\pm 0.03$} &8.57\%\tiny{$\pm 0.01\%$} 
&1.91\tiny{$\pm 0.01$} &3.63\tiny{$\pm 0.01$} &3.02\%\tiny{$\pm 0.06\%$} 
\\
GMSDR & 0.84\tiny{$\pm 0.00$} & 2.63\tiny{$\pm 0.02$} &  43.36\%\tiny{$\pm 0.14\%$} 
& 8.44\tiny{$\pm 0.01$} & 15.89\tiny{$\pm 0.07$} & 47.33\%\tiny{$\pm 0.07\%$} 
&4.77\tiny{$\pm 0.05$} &8.50\tiny{$\pm 0.05$} &8.84\%\tiny{$\pm 0.09\%$} 
&1.94\tiny{$\pm 0.03$} &3.55\tiny{$\pm 0.05$} &3.10\%\tiny{$\pm 0.05\%$} 
\\ 
PDFormer&0.91\tiny{$\pm 0.00$} &2.92\tiny{$\pm 0.02$} &46.57\%\tiny{$\pm 0.23\%$}
&8.62\tiny{$\pm 0.13$} &16.04\tiny{$\pm 0.27$} &48.35\%\tiny{$\pm 0.71\%$}  
& 4.69\tiny{$\pm 0.02$} & 8.60\tiny{$\pm 0.02$} & 8.68\%\tiny{$\pm 0.04\%$} 
&1.87\tiny{$\pm 0.01$}&3.57\tiny{$\pm 0.01$}&2.99\%\tiny{$\pm 0.01\%$} 
\\
\hline
REVIN&1.04\tiny{$\pm 0.01$} &3.42\tiny{$\pm 0.04$} & 53.19\%\tiny{$\pm 0.31\%$} 
&9.96\tiny{$\pm 0.13$} &17.43\tiny{$\pm 0.20$} &55.84\%\tiny{$\pm 0.71\%$}
&	7.24\tiny{$\pm 0.05$} & 11.83\tiny{$\pm 0.04$} & 13.41\%\tiny{$\pm 0.10\%$} 
&3.13\tiny{$\pm 0.02$} &5.87\tiny{$\pm 0.03$} &5.00\%\tiny{$\pm 0.03\%$}\\
PatchTST & 0.95\tiny{$\pm 0.02$}
&3.06\tiny{$\pm 0.09$} &48.60\%\tiny{$\pm 0.98\%$} &
9.22\tiny{$\pm 0.08$}  &17.07\tiny{$\pm 0.15$} &51.71\%\tiny{$\pm 0.44\%$}  & 
5.51\tiny{$\pm 0.13$} & 9.50\tiny{$\pm 0.11$} & 10.20\%\tiny{$\pm 0.24\%$} 
&
2.14\tiny{$\pm 0.02$}&4.08\tiny{$\pm 0.02$} &3.58\%\tiny{$\pm 0.04\%$}  
\\ 
Dlinear&
0.90\tiny{$\pm 0.00$} & 2.81\tiny{$\pm 0.01$} &46.10\%\tiny{$\pm 0.00\%$}
&9.78\tiny{$\pm 0.00$} &17.88\tiny{$\pm 0.00$} &54.85\%\tiny{$\pm 0.01\%$} 
&4.97\tiny{$\pm 0.00$} &9.04\tiny{$\pm 0.21$} & 9.20\%\tiny{$\pm 0.01\%$}  
&2.13\tiny{$\pm 0.00$} &4.11\tiny{$\pm 0.00$}& 3.40\%\tiny{$\pm 0.00\%$}\\ 
OnlineTCN & 0.90\tiny{$\pm 0.00$} & 2.82\tiny{$\pm 0.01$} & 46.35\%\tiny{$\pm 0.11\%$}
& 10.09\tiny{$\pm 0.02$} & 18.20\tiny{$\pm 0.03$} & 56.59\%\tiny{$\pm 0.01\%$}  
&4.78\tiny{$\pm 0.03$} &8.70\tiny{$\pm 0.04$} &9.03\%\tiny{$\pm 0.01\%$}  
&2.08\tiny{$\pm 0.01$} &3.84\tiny{$\pm 0.01$} &3.32\%\tiny{$\pm 0.01\%$} 
\\
FSNet & \underline{0.82}\tiny{$\pm 0.01$} & \underline{2.54}\tiny{$\pm 0.05$} & \underline{42.30\%}\tiny{$\pm 0.57\%$} & 
8.39\tiny{$\pm 0.24$} & 15.45\tiny{$\pm 0.65$} & 46.41\%\tiny{$\pm 1.98\%$} 
&5.79\tiny{$\pm 0.24$}  &11.06\tiny{$\pm 0.24$} &11.06\%\tiny{$\pm 0.44\% $}  
&3.39\tiny{$\pm 0.22$} &5.53\tiny{$\pm 0.40$} &5.41\%\tiny{$\pm 0.35\%$}\\ 
OneNet&
OOM&OOM&OOM&
9.20\tiny{$\pm 0.24$}&16.79\tiny{$\pm 0.48$}&51.40\%\tiny{$\pm 1.33\%$}& 
4.94\tiny{$\pm 0.03$}&8.80\tiny{$\pm 0.06$}&9.14\%\tiny{$\pm 0.06\%$}\% 
&2.00\tiny{$\pm 0.01$}&3.66\tiny{$\pm 0.01$} &3.18\%\tiny{$\pm 0.01\%$} 
\\ 
\rowcolor{background_gray} \name & $\textbf{0.72\tiny{$\pm 0.00$}}^\dagger$ & $\textbf{2.06\tiny{$\pm 0.02$}}^\dagger$ &  $\textbf{36.80\%\tiny{$\pm 0.19\%$}}^\dagger$ & 
$\textbf{7.90\tiny{$\pm 0.02$}}^\dagger$
& $\textbf{14.78\tiny{$\pm 0.02$}}^\dagger$ & $\textbf{44.13\%\tiny{$\pm 0.12\%$}}^\dagger$ & 
\textbf{4.38\tiny{$\pm 0.02$}}$^\dagger$ & \textbf{8.26\tiny{$\pm 0.03$}}$^\ddagger$ & \textbf{8.11\%\tiny{$\pm 0.02\%$}}$^\dagger$ 
&$\textbf{1.67\tiny{$\pm 0.00$}}^\dagger$ &$\textbf{3.25\tiny{$\pm 0.01$}}^\dagger$ &$\textbf{2.67\%\tiny{$\pm 0.01\%$}}^\dagger$ 
\\
\hline
\end{tabular}
\vspace{-0.5em}
\end{table*}

%% file: tables/strategy.tex
\begin{table}[ht]
\vspace{-0.5em}
\centering
\small
\tabcolsep=2.7mm
\caption{
Baseline models with our proposed strategies, i.e., adaptive ST network and AH learning strategy, on the Singapore-T dataset. The * denotes models with the adapter, and + refers to models with both strategies.
}
\vspace{-0.5em}
\label{tab:strategy}
\begin{tabular}{l|c|ccc}
\hline
\textbf{Method} & \textbf{\# Params}& \textbf{MAE}$\downarrow$& \textbf{RMSE}$\downarrow$ & \textbf{WMAPE}$\downarrow$\\
\hline
STGCN & 148K &
8.23\tiny{$\pm 0.08$} & 15.50\tiny{$\pm 0.23$} & 46.17\%\tiny{$\pm 0.46\%$}
\\
STGCN* &  223K 
& 8.13\tiny{$\pm 0.01$}
& 15.27\tiny{$\pm 0.13$}  
& 45.58\%\tiny{$\pm 0.28\%$}
\\
\rowcolor{background_gray} STGCN+ &  223K 
&8.06\tiny{$\pm 0.04$}
& 15.12\tiny{$\pm 0.10$} 
&45.19\%\tiny{$\pm 0.27\%$}
\\
\hline
MTGNN & 233K
&
8.62\tiny{$\pm 0.07$} & 15.70\tiny{$\pm 0.12$} & 47.28\%\tiny{$\pm 0.38\%$}
\\
MTGNN* & 259K 
& 8.13\tiny{$\pm 0.03$}
& 15.26\tiny{$\pm 0.08$}
& 45.56\%\tiny{$\pm 0.19\%$}\\
\rowcolor{background_gray} MTGNN+ &  259K
& 8.03\tiny{$\pm 0.03$}
& 15.05\tiny{$\pm 0.05$}
& 45.05\%\tiny{$\pm 0.15\%$}
\\
\hline
GWNET &  307K 
&  
8.43\tiny{$\pm 0.07$} & 15.70\tiny{$\pm 0.12$} & 47.28\%\tiny{$\pm 0.37\%$}
\\
GWNET*& 332K 
&7.99\tiny{$\pm 0.02$}&14.93\tiny{$\pm 0.07$}&44.83\%\tiny{$\pm 0.14\%$}
\\
\rowcolor{background_gray} GWNET+ &  
332K 
& $\textbf{7.90\tiny{$\pm 0.02$}}$
& $\textbf{14.78\tiny{$\pm 0.02$}}$ & $\textbf{44.13\%\tiny{$\pm 0.12\%$}}$ \\
\hline
\end{tabular}
\vspace{-1em}
\end{table}

%% file: tables/test_mem.tex
\begin{table}[!ht]
\vspace{-1em}
\begin{center}
\caption{Inference time and memory usage comparisons on Singapore-T dataset.
}
\vspace{-0.5em}
\label{tab:speed_mem_comparison}
\small
\tabcolsep=0.6mm
\begin{tabular}{l|ccc|c}
\hline
\multirow{2}{*}{\textbf{Method}} & \textbf{Total Inference} & \textbf{Inference Time} & \multirow{2}{*}{\textbf{Speed-up}}& \textbf{Memory}\\
& \textbf{Time (s)}&\textbf{/Sample (s)}&&\textbf{Usage (MB)} \\
\hline
STGCN
&34.6954 & 0.0026 & 57.28$\times$&
1428\\
GWNET
& 198.4679 & 0.0152 & 10.01$\times$
&1566	\\
AGCRN & 
312.9258 & 0.0239& 6.35$\times$& 
1580  \\
MTGNN 
&108.8445 &0.0083&18.26$\times$&
1466	\\
GMSDR 
&981.2985 & 0.0749 & 2.03$\times$
&1218  \\
PDFormer
&220.3777&0.0168 & 9.02$\times$&
1432  \\
\hline
REVIN
&51.1791& 0.0039 & 38.83$\times$
&1400\\
PatchTST
&97.5924& 0.0075 & 20.37$\times$
&1190	\\
Dlinear
&190.3008&0.0145&10.45$\times$
&1130  \\
OnlineTCN
&356.2887&0.0272&5.58$\times$
&1634 \\
FSNet
&1821.9164&0.1392&1.09$\times$
&1556	\\
OneNet
&1987.6175 & 0.1518 &$1\times$&
1564\\
\rowcolor{background_gray} \name
& 1193.7509 & 0.0912 &1.66$\times$ &
1823 \\ 
\hline
\end{tabular}
\end{center}
\vspace{-1em}
\end{table}

%% file: figures/exp/ablation.tex
\begin{figure}[!t]
\vspace{-1em}
  \centering
\includegraphics[width=1.0\linewidth]{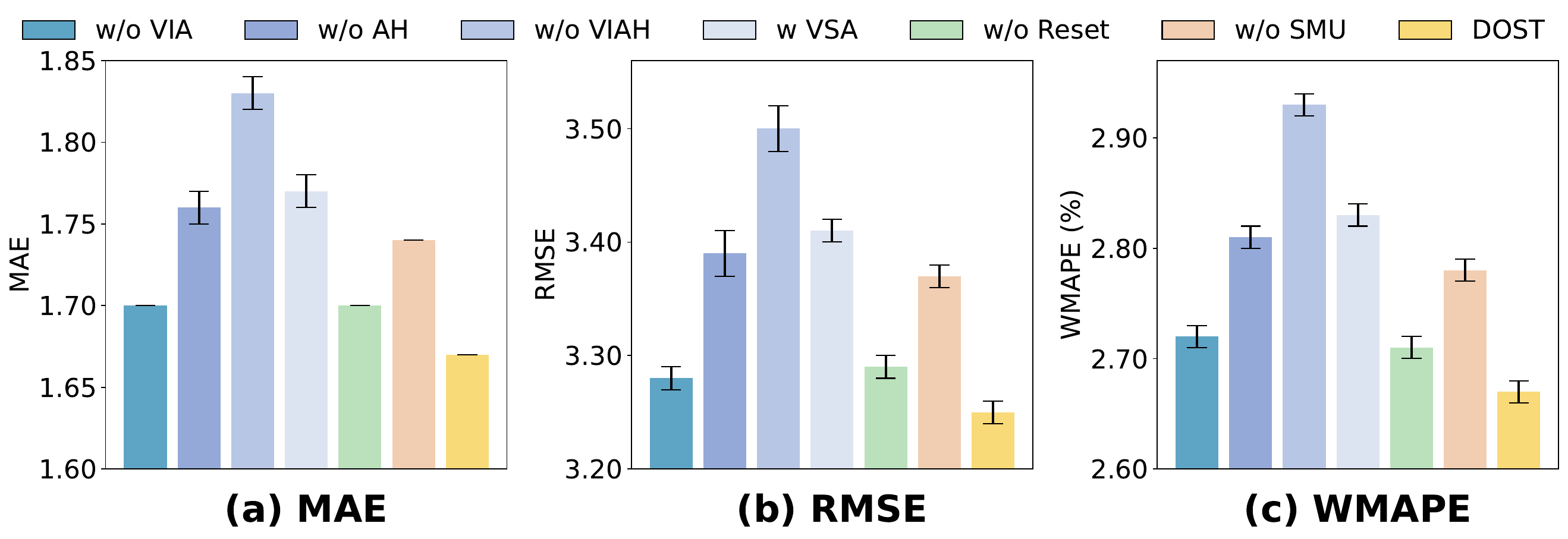}
\vspace{-1.5em}
\caption{
Ablation study of DOST on PEMS-BAY dataset.
}
\vspace{-1em}
\label{fig:ablation}
\end{figure}

%% file: tables/onlinestrategy.tex
\begin{table}[ht]
\vspace{-0.5em}
\centering
\small
\tabcolsep=3.2mm
\caption{
Model update strategy study on PEMS-BAY dataset.
}
\vspace{-0.5em}
\label{tab:online_update}
\begin{tabular}{l|ccc|c}
\hline
\textbf{Method} & \textbf{MAE}$\downarrow$& \textbf{RMSE}$\downarrow$ & \textbf{WMAPE}$\downarrow$& \textbf{Time (s)}\\
\hline
w/o H & 1.69\tiny{$\pm 0.01$}
& 3.28\tiny{$\pm 0.01$}
 & 2.70\%\tiny{$\pm 0.01\%$} & 14377.68
\\
w ER & 1.70\tiny{$\pm 0.01$}
& 3.30\tiny{$\pm 0.01$}
& 2.71\%\tiny{$\pm 0.01\%$}
&  22002.96
\\
w ERH &  1.68\tiny{$\pm 0.01$} 
&  3.26\tiny{$\pm 0.01$}
& 2.68\%\tiny{$\pm 0.01\%$} 
& 9436.99
\\
w SMUR &  1.66\tiny{$\pm 0.00$} 
& 3.23\tiny{$\pm 0.01$}
&2.65\%\tiny{$\pm 0.01\%$} &9215.48
\\
Full &1.66\tiny{$\pm 0.00$}&3.23\tiny{$\pm 0.01$}&2.65\%\tiny{$\pm 0.01\%$}&9015.64\\
\rowcolor{background_gray} \name & 1.67\tiny{$\pm 0.00$} &3.25\tiny{$\pm 0.01$} &2.67\%\tiny{$\pm 0.01\%$}&8131.00\\
\hline
\end{tabular}
\end{table}

%% file: figures/exp/hyperparameter.tex
\begin{figure}[h]
  \centering
\includegraphics[width=1.0\linewidth]{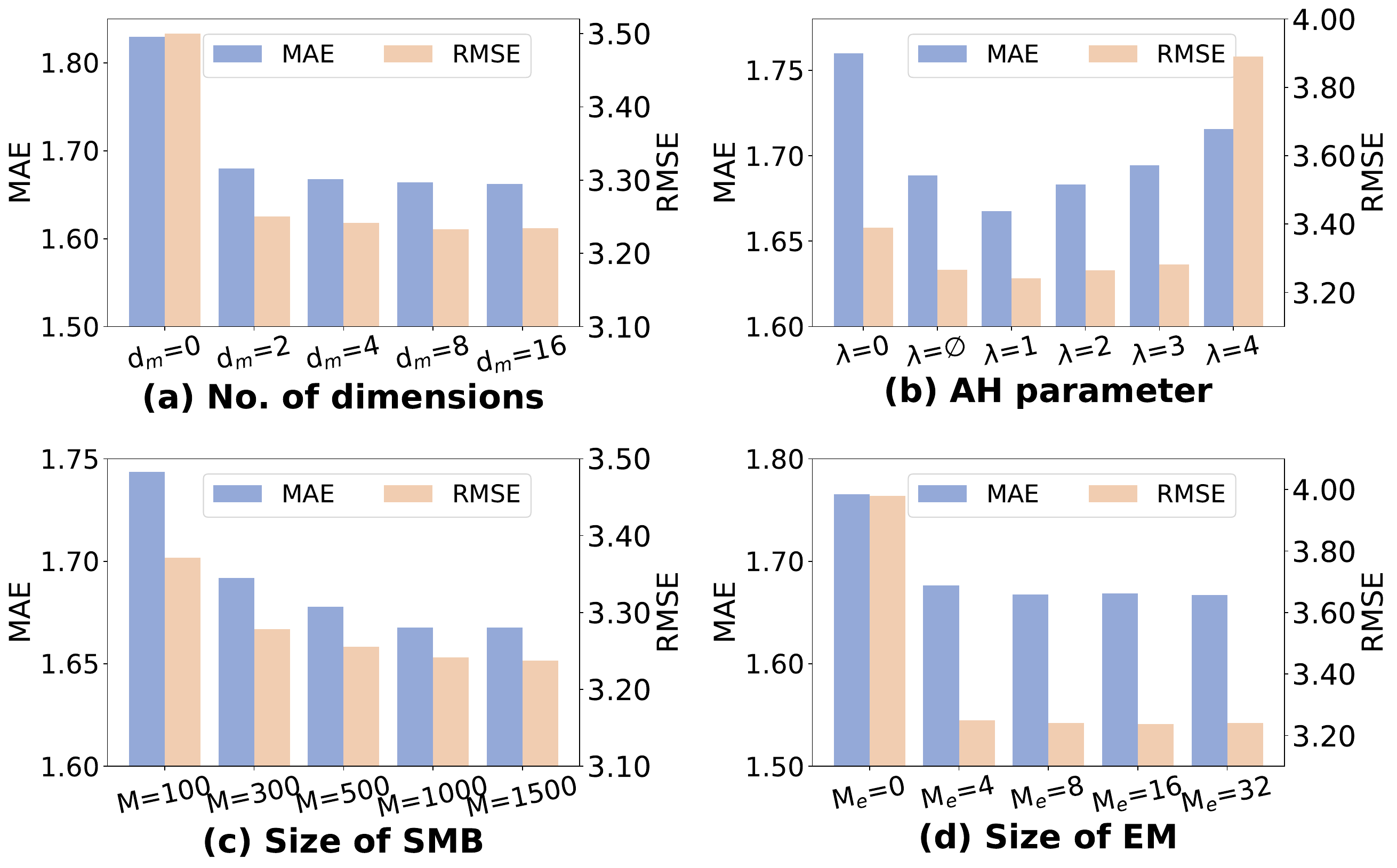}
\caption{Effects of hyperparameters on PEMS-BAY dataset.}
\label{fig:hyperparameter}
\end{figure}

%% file: sections/5.relatedwork.tex
\section{Related Work}

\vspace{0.5em}
\noindent
\textbf{Urban Spatio-Temporal Forecasting} is a crucial task in smart city development, serving numerous urban applications and attracting notable attention.
Early attempts employed traditional time series models, such as Autoregressive Integrated Moving Average (ARIMA)~\cite{williams2003modeling, tran2015multiplicative}, Holt-Winters methods~\cite{de2013holt, daraghmi2015improved}, and Support Vector Regression (SVR)~\cite{wei2013adaptive,hong2006highway}. However, these methods often fail to capture the complex ST correlations in urban data. In recent decades, deep learning-based approaches, especially Spatio-Temporal Graph Neural Networks (STGNNs), have gained prominence in this task. They typically integrate Graph Convolutional Networks (GCN)~\cite{kipf2017semi,zheng2020gman,zhou2020reinforced} or attentions~\cite{zhou2020spatiotemporal, zhao2023spatio} with various sequence models, such as Recurrent Neural Networks (RNNs)~\cite{hochreiter1997long,gao2016deep}, Convolutional Neural Networks (CNNs)~\cite{wang2022periodic, graphwvenet}, and attentions~\cite{vaswani2017attention,liu2023spatio}.
These methods effectively capture complex ST patterns, but most focus on dynamic spatial correlations using dynamic graphs~\cite{yu2023towards} and often overlook temporal dynamics, where data arrives sequentially and distributions shift in real-world online deployment. Similarly, recent graph-free models employing normalization~\cite{deng2021st} or identity embedding~\cite{shao2022spatial, liu2023spatio} also neglect these temporal dynamics.
Some studies~\cite{zhang2021dac, zhang2022urban} incorporate continual learning for urban ST forecasting, but they are limited to grid-based ST data using CNNs, which are not suitable for various non-Euclidean urban ST data types, such as traffic speed.
Recent works~\cite{zhou2023maintaining,wang2024citycan} leverage causal theory~\cite{pearl2000models} to uncover invariant relationships in training data, aiming to address distribution shifts between training and testing phases. Additionally, some methods~\cite{kim2021reversible, nie2022time} employ learnable normalization trained on fixed samples to tackle the out-of-distribution problem.
However, these methods still face challenges in real-world applications, where certain invariant relationships may change over extended periods, especially during unprecedented events like the COVID-19 pandemic~\cite{cruz2021impact}.
Some concurrent studies~\cite{miao2024unified, li2024flashst} adopt continual learning for ST data but neglect location-specific distribution shifts.
In this work, we utilize online continual learning to address gradual, location-specific shifts, which is orthogonal to the learning of invariant relationships and learnable normalization.

\vspace{0.5em}
\noindent
\textbf{Online Continual Learning} focuses on adapting to shifts in data distribution and has proven highly effective in various tasks, including image classification~\cite{buzzega2020dark,harun2023siesta,gunasekara2023survey}, semantic segmentation~\cite{douillard2021plop, cermelli2022incremental}, and natural language processing~\cite{houlsby2019parameter,pfeiffer2020adapterhub}. 
Recent research has introduced architecture-based approaches for online time series forecasting. 
For example, FSNet~\cite{pham2022learning} incorporates additional memory mechanisms to adapt to new data, and OneNet~\cite{zhang2023onenet} employs reinforcement learning to adjust weights for two separate forecasters.
These models are designed for data with unstable temporal patterns~\cite{shao2023exploring} and often employ full fine-tuning strategies.
However, urban ST data usually exhibits more stable temporal patterns, with data distribution shifts over longer periods, thereby reducing the need for frequent full fine-tuning.
In recent years, adapter-based frameworks~\cite{houlsby2019parameter, pfeiffer2020mad, zhou2023one,zhang2023llama} have shown remarkable effectiveness in handling unseen tasks. For example, Adapter-BERT \cite{houlsby2019parameter} adds a two-layer feed-forward network and fine-tunes these layers for unseen tasks. MAD-X \cite{pfeiffer2020mad} introduces an invertible adapter to capture token-level language-specific transformations.
Additionally, recent work~\cite{harun2023siesta} demonstrates that a sleep mechanism can achieve efficient online image classification.
Inspired by these works, we introduce a variable-independent adapter to address location-specific shifts and propose a novel AH learning strategy with a streaming memory update to reduce retraining frequency while maintaining forecasting performance.

%% file: sections/6.conclusion.tex
\section{Conclusions}
\label{sec:conclusion}
In this paper, we first investigate the gradual distribution drift characteristic of urban sequential spatio-temporal (ST) data, and then introduce \name, a novel distribution-aware online continual learning for urban ST forecasting. \name contains two key novelties - an \textit{adaptive ST network} and an \textit{awake-hibernate (AH) learning strategy}.
These elements can be seamlessly integrated into existing offline ST networks to boost their performance.
The variable-independent adapter within the adaptive ST network enable \name to adapt to the varied distribution shifts across different urban locations.
The AH learning strategy efficiently reduces computational overhead by intermittently hibernating network updates and mitigates catastrophic forgetting through a streaming memory update mechanism.
Extensive experimental results and analyses conducted on four real-world datasets validate the effectiveness of \name.